%% file: 1_main.tex
\documentclass[letterpaper, 10 pt, conference]{ieeeconf}  

\IEEEoverridecommandlockouts                              

\input{_format}
\title{\LARGE \bf
 DOBB: Dynamic Oriented Bounding Boxes for Multi-Agent Collision Avoidance under State Uncertainty and Heterogeneous Dynamics 
}

\author{Pedro Antonio Alarcon Granadeno$^{1}$ and Jane Cleland-Huang$^{2}$
\thanks{*This work was supported by the USA National Science Foundation under Grant 1931962}
\thanks{$^{1}$Pedro Antonio Alarcon Granadeno is a PhD student in the Computer Science and Engineering department at the University of Notre Dame, IN, USA
        {\tt\small palarcon@nd.edu}}%
\thanks{$^{2}$Jane Cleland-Huang is a Professor in the Department of Computer Science and Engineering, University of Notre Dame, IN, USA     {\tt\small JaneHuang@nd.edu}}%
}

\begin{document}
\maketitle
\thispagestyle{empty}
\pagestyle{empty}

\begin{abstract}
Learning-based decentralized collision avoidance must remain reliable under state-estimation uncertainty and heterogeneous agent dynamics. Existing methods typically address these factors through training-domain variation or by exposing uncertainty and capability estimates to the policy at runtime. However, both leave the policy to infer how these quantities should alter the physical separation required for safe interaction, which can couple collision-avoidance behavior to relationships learned from the training distribution. We introduce the Dynamic Oriented Bounding Box (DOBB), an online geometric representation that maps the current state estimate, uncertainty, and agent-specific response characteristics into a dynamic, direction-dependent safety margin. We derive geometric features that jointly characterize the relative configuration and severity of interactions between agents’ DOBBs, along with a reward function that preserves the physical interpretation of those interactions. Across four evaluation environments, DOBB achieves the lowest mean total failure rate of 2.6\%. This performance persists under out-of-distribution shifts in actuation dynamics and state-estimation uncertainty, with collision-based failure remaining at or below 2.3\% in every tested condition. We attribute this robustness to the geometric abstraction of the safety margin, which is less coupled to the specific uncertainty and actuation regimes encountered during training. Code open-sourced at: \href{https://anonymous.4open.science/r/DOBB-Collision-Avoidance-8D16/README.md}{https://anonymous.4open.science/r/DOBB-Collision-Avoidance-8D16/README.md}
\end{abstract}

\input{2_Introduction}

\input{3_Related_Work}

\input{4_Problem_Setup}
\input{5_Methodology}

\input{6_Experimental_Setup}
\input{7_Evaluation}
\input{8_Conclusion}

\bibliographystyle{IEEEtran}
\bibliography{_bibby}

\end{document}

%% file: _format.tex
\usepackage{amsmath} 
\usepackage{amssymb}  
\let\labelindent\relax
\usepackage[inline]{enumitem}
\usepackage{hyperref}
\usepackage{xcolor} 
\usepackage{array}
\usepackage{graphicx} 
\usepackage{booktabs}  
\usepackage{caption}   
\usepackage{lipsum} 
\usepackage{multirow}
\usepackage{booktabs} 
\usepackage{subcaption}
\usepackage{hyperref}
\hypersetup{
    colorlinks=true,
    linkcolor=blue,
    filecolor=magenta,      
    urlcolor=cyan,
    pdfpagemode=FullScreen,
    }

\usepackage{tcolorbox}

\usepackage{amsthm}

\usepackage{graphicx}
\usepackage{adjustbox}
\usepackage{lipsum}
\usepackage{tabularx}

\usepackage{adjustbox}
\usepackage{subcaption}
\usepackage{caption}

%% file: 2_Introduction.tex
\section{Introduction}
\label{sec:introduction}

Learning-based decentralized collision avoidance (DCA) enables multi-agent navigation from locally available observations without centralized coordination. In physical fleets, the safety of an interaction depends not only on the estimated relative configuration, but also on uncertainty in that estimate and differences in agent physical footprint and actuation response. State-estimation uncertainty and agent heterogeneity have therefore become recurring concerns in learning-based DCA.

Two recurring themes characterize much of the learning-based DCA literature. First, and most commonly, \textbf{experience-based approaches} accommodate uncertainty and heterogeneity through the observations, interactions, and temporal context available to the policy during training. Examples include noise-augmented observations, variation in agent(s) properties (i.e. radius, swarm size) or behavior (cooperative vs. non-cooperative), and temporal context from stacked observations fed to recurrent encoders \cite{long2018towards, wang2020two, everett2021collision, thumiger2021multi}. These mechanisms improve robustness by emulating real-life conditions and interactions during learning and/or leverage recurrent architectures for latent-state inference. Second, more recently with a smaller body of work,  \textbf{explicit-runtime approaches} quantify uncertainty or related predictive confidence during execution and incorporate it directly into the decision process. Depending on the formulation, the resulting estimate may enter the policy observation or reward, modify an optimization objective or action distribution, or trigger a fallback controller \cite{lutjens2019safe, fan2020learning, golchoubian2024uncertainty, flogel2025disentangling}. 

These two approaches impose different inference burdens on the collision-avoidance policy. Consider an experience-based policy trained with noise-augmented position measurements to emulate localization uncertainty. At execution, the policy observes the corrupted measurements but is not given the underlying noise magnitude; it must infer the severity of uncertainty from the observed fluctuations and subsequently learn how that uncertainty should alter its avoidance behavior. An explicit-runtime formulation can partially alleviate this burden by quantifying an estimate of this uncertainty (e.g. position covariance) at each time step. Explicit access to uncertainty therefore removes the need to infer its severity, but it does not specify how uncertainty and agent heterogeneity should alter the separation required for safe interaction. 

This limitation can be viewed as a representation problem. In collision avoidance, state uncertainty and agent response characteristics affect the spatial margin that should be maintained during an interaction. Consequently, the same estimated relative configuration can imply different separation requirements across uncertainty levels and agent capabilities, where these effects may be directionally dependent. For example, identical estimated positions and velocities can correspond to materially different interactions when one agent is poorly localized and has tighter acceleration limits (e.g., due to a lower thrust-to-weight ratio), whereas another is accurately localized and more maneuverable. When these quantities are provided only as policy inputs, their spatial implications must still be learned from experience. This learned mapping can become coupled to the uncertainty and actuation regimes represented during training, making the resulting collision-avoidance behavior susceptible to degradation when deployment conditions depart from that distribution.

We address this representation problem with the Dynamic Oriented Bounding Box (DOBB), an online geometric abstraction that maps the current state estimate, its uncertainty, and agent-specific response characteristics into a dynamic, direction-dependent safety margin. Rather than exposing these quantities independently to the policy, DOBB represents their joint effect on the spatial relationship between interacting agents and recomputes this representation as operating conditions change. We further derive geometric interaction features that characterize the relative configuration and severity of interactions between agents' DOBBs, along with a reward that preserves the physical interpretation of these interactions. The evaluation in Sec. \ref{sec:evaluation} shows that DOBB reduces collision failures relative to both experience-based and explicit-runtime representations, with the improvement persisting under out-of-distribution shifts in actuation dynamics and state-estimation uncertainty.

%% file: 3_Related_Work.tex
\section{Related Work}
\label{sec:related-work}

Experience-based approaches seek robustness by embedding uncertainty and agent variation in the conditions encountered during learning or by providing additional context from which partially observed state can be inferred. In \cite{wang2020two}, observed neighbor positions and velocities are perturbed with Gaussian noise sampled across a range of magnitudes. The resulting blind-noise policy handles unknown noise magnitudes within that range, but degrades when evaluated under a different noise distribution. In \cite{everett2021collision} and \cite{thumiger2021multi}, policies are trained across varying agent populations and behaviors, with LSTM encoders used to represent a variable number of surrounding agents. \cite{everett2021collision} additionally varies radius and preferred speed, while \cite{thumiger2021multi} injects sensing and action noise. The latter reports a simulation-to-real gap attributed in part to under-actuated quadrotor dynamics absent from training. In \cite{long2018towards}, three consecutive LiDAR scans provide short-horizon temporal information from which the policy can infer obstacle motion and geometry. Although trained with homogeneous discs, the resulting policy generalizes to previously unseen square robots. These studies demonstrate that broader training experience can improve robustness and generalization. However, when uncertainty magnitude or response characteristics are not directly represented, the policy must infer them from the available evidence and learn how their severity should alter avoidance behavior. This reliance on learned relationships can degrade when deployment conditions depart from those represented during training.

Explicit-runtime approaches reduce part of this inference burden by quantifying uncertainty or related predictive confidence during execution and allowing the current estimate to influence action selection directly. In \cite{lutjens2019safe}, bootstrapped LSTM ensembles and MC dropout estimate the mean and variance of collision probability for candidate motion primitives, and an MPC penalizes both expected collision risk and model uncertainty. The resulting uncertainty signal identifies novel observations and reduces collisions under sensing perturbations absent from training. In \cite{fan2020learning}, a GRU predicts the next LiDAR scan and its per-beam aleatoric uncertainty, which is mapped to the variance of a SAC policy so that higher uncertainty elicits more conservative behavior. In \cite{golchoubian2024uncertainty}, future pedestrian positions are represented as bivariate Gaussians, with the predicted covariance incorporated into both the policy state and a collision-probability reward. This reduces collisions relative to the corresponding uncertainty-unaware prediction model. In \cite{flogel2025disentangling}, uncertainty in the policy distribution is combined with proximity conditions to trigger an ORCA fallback, where surrounding human radii are enlarged by a fixed factor. These approaches make uncertainty available during execution but do not derive a state-dependent safety margin from it. In \cite{lutjens2019safe,fan2020learning,golchoubian2024uncertainty}, uncertainty changes the optimization objective, policy distribution, or learning signal, while the physical clearance that results is still determined by optimization or policy learning. \cite{flogel2025disentangling} comes closest to an explicit geometric response, but only after a manually calibrated high-risk condition is satisfied. The controller then reverts to ORCA with a fixed 1.5-times radius enlargement. The resulting inflation is isotropic, discontinuous, and independent of further changes in uncertainty once the threshold has been exceeded.

A complementary line of work modifies collision geometry directly. Prior ORCA variants inflate the represented agent radius to account for differential-drive constraints \cite{5652073}, tracking-error bounds \cite{alonso2013optimal}, or observation noise \cite{wang2020two}. These methods alter the safety margin explicitly, but the added clearance remains isotropic and follows a prescribed rule for the corresponding condition. Such inflation can consume available free space unnecessarily when the source of additional risk is directional. Work in \cite{zhu2022collision} address this limitation with an Oriented Bounding Capsule for shape heterogeneity and a speed-dependent risk capsule whose forward length increases by \(\|v_i\|\delta t\). This representation recovers lateral free space discarded by circular models and improves performance in dense heterogeneous scenes. \cite{zhu2022collision} therefore provide the closest geometric precursor to DOBB. Unlike DOBB, however, \cite{zhu2022collision} assumes perfect sensing and instantaneous velocity execution.

Prior work addresses complementary aspects of the same representation problem. Learning-based approaches either broaden the conditions encountered during training or expose uncertainty and capability estimates explicitly at runtime, while leaving their spatial implications largely to policy learning. Geometric approaches only partially address this mapping: existing methods typically use prescribed, often isotropic margin modifications for individual effects such as sensing noise, tracking error, or kinematic constraints, while the closest learning-based geometric treatment \cite{zhu2022collision} captures shape and speed heterogeneity but does not account for state-estimation uncertainty or finite actuation response. DOBB instead provides a unified geometric representation in which state uncertainty and heterogeneous actuation response jointly determine a dynamic, direction-dependent safety margin exposed to the policy.

%% file: 4_Problem_Setup.tex
\section{Problem Setup}
\label{sec:problem-setup}
\subsection{Decentralized Navigation Problem}
\label{subsec:dec-navigation-prob}
We consider planar decentralized navigation among $N$ agents, indexed by $i\in\{1,\ldots,N\}$, each assigned a fixed goal $g_i\in\mathbb{R}^2$. Agent heterogeneity is represented through physical footprint, preferred speed, and actuation-response limits that govern how rapidly commanded motion can be realized. We collect these properties as
\begin{equation}
\eta_i =
\left(
r_i,\,
v_i^{\mathrm{pref}},\,
a_i^{\max},\,
j_i^{\max},\,
\omega_i^{\max},\,
\dot{\omega}_i^{\max},\,
\tau_i
\right),
\label{eq:agent_heterogeneity}
\end{equation}
where $r_i$ and $v_i^{\mathrm{pref}}$ denote the physical radius and preferred speed of agent $i$; $a_i^{\max}$, $j_i^{\max}$, $\omega_i^{\max}$, $\dot{\omega}_i^{\max}$ and $\tau_i$ are its acceleration, jerk, yaw-rate, yaw-acceleration limits, and response-latency allowance, respectively. 

Let $t\in\mathbb{Z}_{\geq 0}$ index policy decision steps separated
by the interval $\Delta t$, and let $x_i^t$ denote the physical state
of agent $i$ at time $t\Delta t$:
\begin{equation}
x_i^t =
\left(
p_i^t,\,
s_i^t,\,
\psi_i^t,\,
\alpha_i^t,\,
\omega_i^t
\right),
\label{eq:agent_state}
\end{equation}
where $p_i^t$, $s_i^t$, $\psi_i^t$, $\alpha_i^t$, and $\omega_i^t$ denote position, speed, heading, longitudinal acceleration, and yaw rate. At each decision step, the policy selects the target speed and relative heading command $u_i^t=(v_{i,\mathrm{cmd}}^t,\Delta\psi_{i,\mathrm{cmd}}^t)$. The state then evolves as $x_i^{t+1}=f(x_i^t,u_i^t,\eta_i,\Delta t)$, where $f$ is defined by the constrained motion model in Sec.~\ref{subsec:actuation-constrained-motion}

Each agent seeks to reach its goal while avoiding physical collisions. Goal arrival occurs when its position lies within $\epsilon_g$ of the goal, $|p_i^t-g_i|_2\leq\epsilon_g$, while collision occurs when two agents' circular footprints touch or overlap, $|p_i^t-p_j^t|_2\leq r_i+r_j$. An agent terminates upon goal arrival, collision, or timeout. The physical state $x_i^t$ is maintained as ground truth for propagating the dynamics and evaluating these conditions.

The policy does not observe $x_i^t$ directly. Physical state must instead be reconstructed from imperfect onboard sensing, external localization, or communicated state estimates, and is therefore subject to estimation error. A state estimator provides agent $i$ with its own estimate $\hat{x}_i^t$ and estimates $\hat{x}_j^t$ of locally observed neighbors $j\in\mathcal{N}_i^t$. Each estimate is associated with an uncertainty description, denoted $\Sigma_i^t$ and $\Sigma_j^t$, respectively. Because action selection is based only on locally available estimated state, we formulate the problem as a decentralized partially observable Markov decision process. All learning agents share policy parameters $\theta$ and select $u_i^t\sim\pi_\theta(\cdot\mid o_i^t)$, where $o_i^t$ contains the locally available information. 

\subsection{Actuation-Constrained Motion}
\label{subsec:actuation-constrained-motion}
For simplicity, many learning-based DCA formulations assume instantaneous motion, such that policy commands are realized immediately. However, a policy trained under this assumption is susceptible to sharp degradation when deployed on an agent with finite actuation response. In fact, \cite{thumiger2021multi} identifies under-modeled actuation as a contributing factor to simulation-to-real degradation, and our evaluation corroborates a similar conclusion (Sec.~\ref{subsec:rq1}). We therefore treat each policy output as a target command that is realized subject to the agent-specific translational and rotational response limits in $\eta_i$.

Using the state defined in Eq.~\ref{eq:agent_state}, the motion of agent $i$ follows
\begin{equation}
\dot{x}_i =
\left[
s_i\cos\psi_i \;
s_i\sin\psi_i \;
\alpha_i \;
\omega_i \;
j_i \;
\beta_i
\right]^\top
\label{eq:agent_dynamics}
\end{equation}
subject to
\begin{equation}
|\alpha_i| \leq a_i^{\max},\;
|j_i| \leq j_i^{\max},\;
|\omega_i| \leq \omega_i^{\max},\;
|\beta_i| \leq \dot{\omega}_i^{\max},
\label{eq:actuation_constraints}
\end{equation}
where $j_i$ and $\beta_i$ denote longitudinal jerk and yaw acceleration. The command $u_i^t$ is held over the interval
$[t\Delta t,(t+1)\Delta t]$, while the longitudinal and rotational
responses evolve toward the commanded targets subject to the limits in
Eq.~\eqref{eq:actuation_constraints}. Integrating the constrained
dynamics over this interval defines the state transition
\begin{equation}
x_i^{t+1}
=
x_i^t
+
\int_{t\Delta t}^{(t+1)\Delta t}
\dot{x}_i(\xi)\,d\xi
=
f\!\left(x_i^t,u_i^t,\eta_i,\Delta t\right).
\label{eq:constrained_state_transition}
\end{equation}
Identical policy commands can therefore produce different trajectories across agents when their response limits differ.

%% file: 5_Methodology.tex
\input{artifacts/fig_dobb_construction}

\section{Methodology}
\label{sec:methodology}

At decision step $t$, before selecting $u_i^t$, DOBB maps the estimated state $\hat{x}_i^t$, uncertainty $\Sigma_i^t$, heterogeneous properties $\eta_i$, and most recently applied command $u_i^{t-1}$ to three nested oriented rectangles, $B_i^{\mathrm{occ},t}\subseteq B_i^{\mathrm{proj},t}\subseteq B_i^{\mathrm{res},t}$. For $z\in\mathcal{Z}=\{\mathrm{occ},\mathrm{proj},\mathrm{res}\}$, each layer is anchored at the current estimated pose and parameterized by body-frame extents $\ell_i^{z,t}=[\ell_{i,f}^{z,t},\,\ell_{i,b}^{z,t},\,\ell_{i,l}^{z,t},\,\ell_{i,r}^{z,t}]^\top$. We write $B(p,\psi,\ell)$ for the rectangle defined by extents $\ell$, rotated by $\psi$ and translated to $p$, such that $B_i^{z,t}=B(\hat p_i^t,\hat\psi_i^t,\ell_i^{z,t})$. The three layers differ only in how these extents are constructed, as shown in Fig.~\ref{fig:dobb_construction}.

\input{5a_DOBB}

\input{5b_Observations}

\input{5c_Reward}

%% file: artifacts/fig_dobb_construction.tex
\begin{figure*}[t]
    \centering
    \subfloat[Occupancy $B_i^{\mathrm{occ},t}$]{
        \includegraphics[width=0.24\textwidth]
        {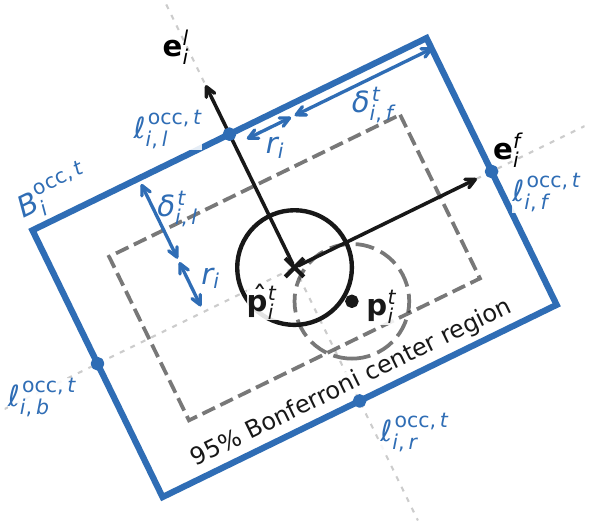}
        \label{fig:dobb_occupancy}
    }%
    \subfloat[Projected motion $B_i^{\mathrm{proj},t}$]{
        \includegraphics[width=0.24\textwidth]
        {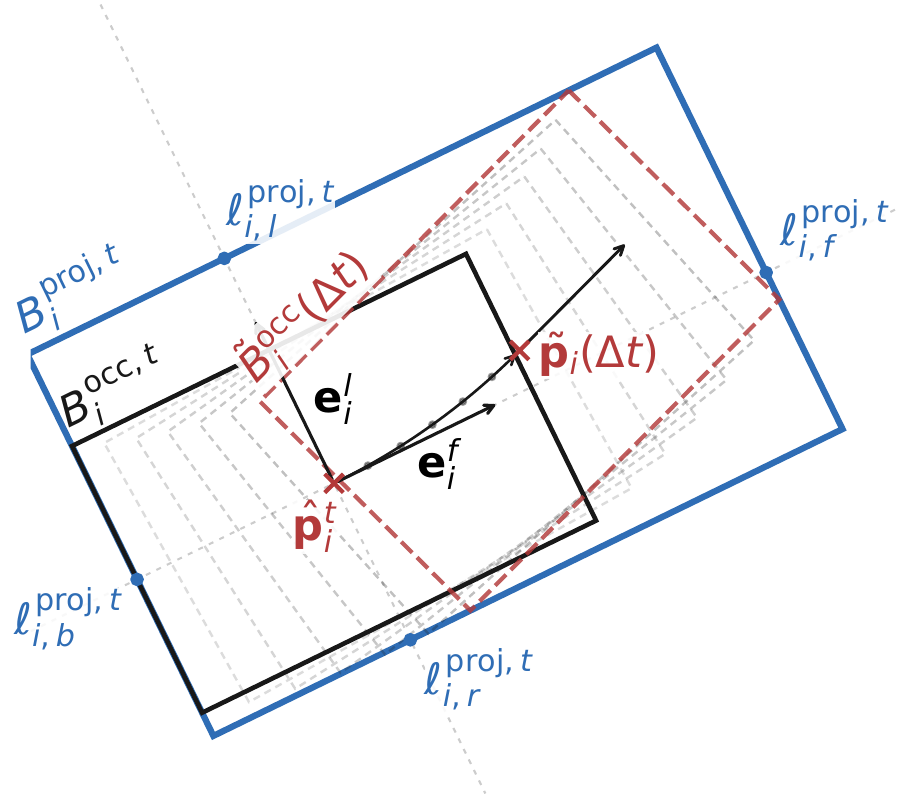}
        \label{fig:dobb_projected}
    }%
    \subfloat[Stopping reservation $B_i^{\mathrm{res},t}$]{
        \includegraphics[width=0.24\textwidth]
        {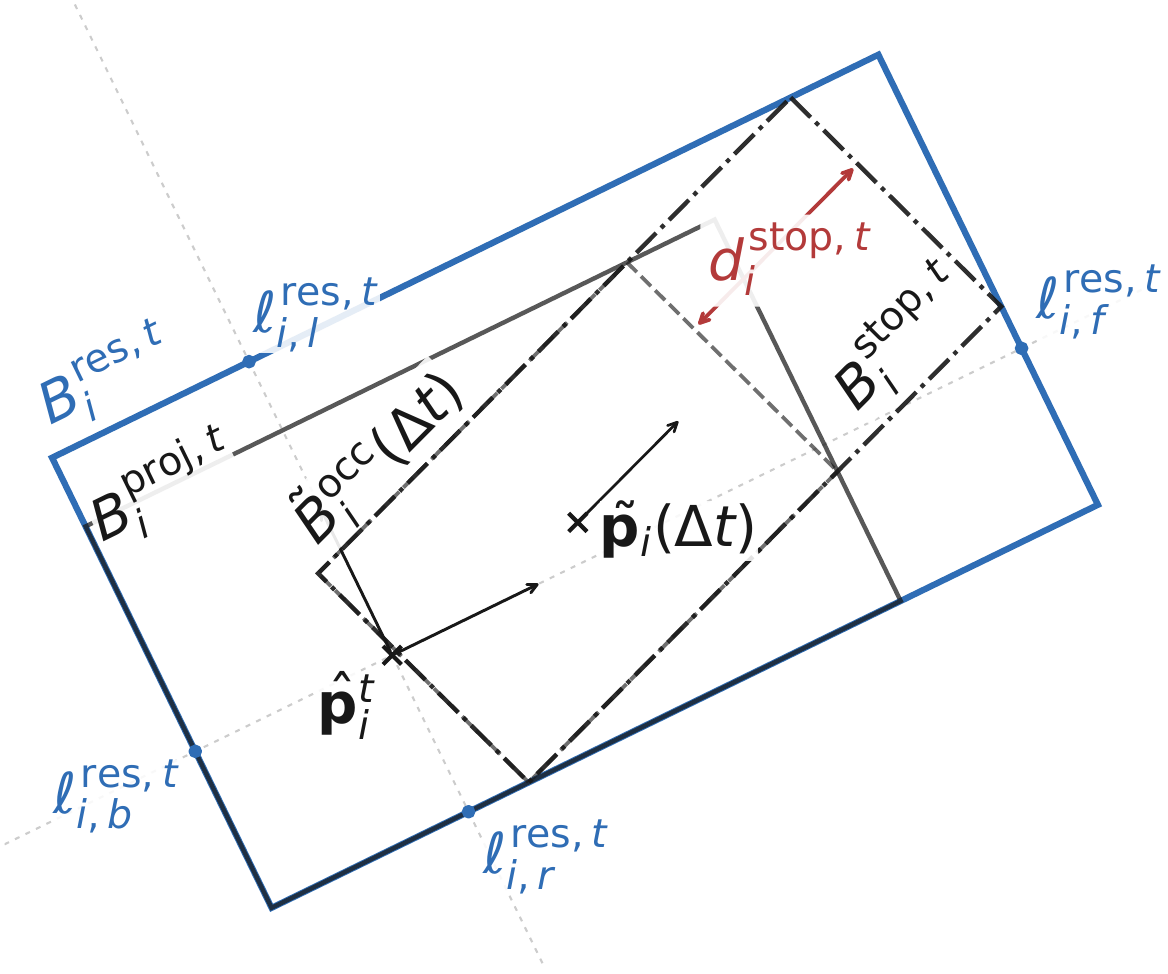}
        \label{fig:dobb_stopping}
    }%
    \subfloat[Final layered DOBB]{
        \includegraphics[width=0.24\textwidth]
        {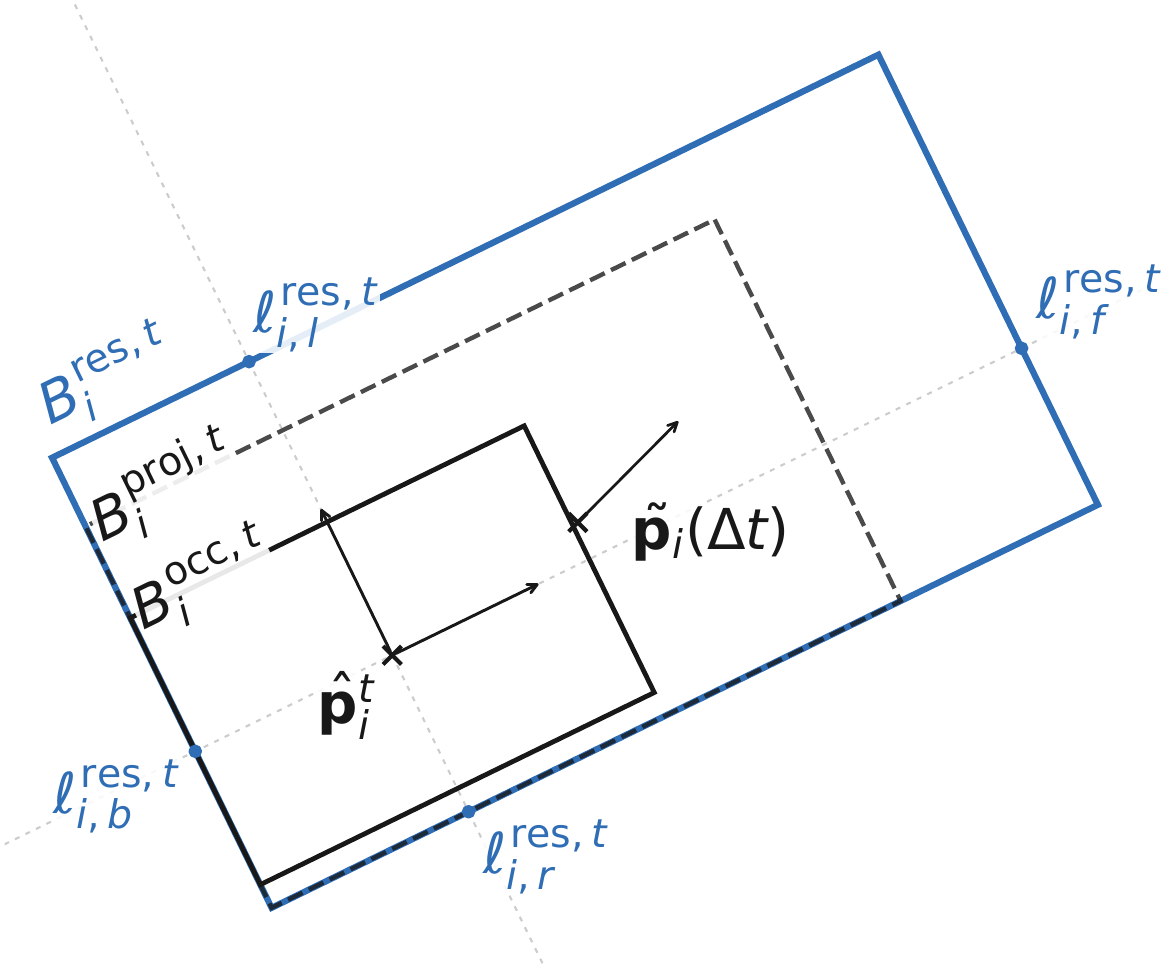}
        \label{fig:dobb_final}
    }

    \caption{Sequential construction of the DOBB geometry at decision step $t$.}
    \label{fig:dobb_construction}
    \vspace{-12pt}
\end{figure*}

%% file: 5a_DOBB.tex
\subsection{DOBB Layer Construction}
\label{subsec:dobb}
\subsubsection{Occupancy Layer}

The occupancy layer incorporates uncertainty in the estimated position into the agent’s physical footprint. Let $\Sigma_{p,i}^t$ denote the positional covariance block of $\Sigma_i^t$, and let $e_i^f$ and $e_i^l$ denote the forward and left body axes at the estimated heading $\hat{\psi}_i^t$. Using simultaneous 95\% Bonferroni-corrected Gaussian bounds, where $\gamma\approx2.241$, the body-axis uncertainty bounds are

\[
\delta_{i,f}^t
=
\gamma
\sqrt{(e_i^f)^\top\Sigma_{p,i}^t e_i^f},
\qquad
\delta_{i,l}^t
=
\gamma
\sqrt{(e_i^l)^\top\Sigma_{p,i}^t e_i^l}.
\]

Adding these bounds to the physical radius gives

\begin{equation}
\ell_i^{\mathrm{occ},t}
=
\left[
r_i+\delta_{i,f}^t,\;
r_i+\delta_{i,f}^t,\;
r_i+\delta_{i,l}^t,\;
r_i+\delta_{i,l}^t
\right]^\top .
\label{eq:occupancy_extents}
\end{equation}

The resulting occupancy layer therefore bounds the physical footprint using the positional uncertainty projected along the current body axes, as shown in Fig.~\ref{fig:dobb_occupancy}.

\subsubsection{Projected-Motion Layer}

The occupancy layer captures the agent's current uncertainty-bounded footprint but not the additional space associated with its near-term motion. We therefore form a one-step continuation prediction under the most recently applied command $u_i^{t-1}$ by propagating the estimated state for one decision interval using the actuation-constrained transition $f$ in Eq.~\ref{eq:constrained_state_transition}. The terminal projected state is
\begin{equation}
\tilde{x}_i(\Delta t)
=
f\left(
\hat{x}_i^t,
u_i^{t-1},
\eta_i,
\Delta t
\right).
\label{eq:projected_terminal_state}
\end{equation}

As shown in Fig.~\ref{fig:dobb_projected}, the occupancy footprint is then anchored at the projected pose, which we denote by $\tilde B_i^{\mathrm{occ}}(\Delta t) = B(\tilde p_i(\Delta t),\tilde\psi_i(\Delta t), \ell_i^{\mathrm{occ},t})$. We express the current occupancy $B_i^{\mathrm{occ},t}$ and projected occupancy $\tilde B_i^{\mathrm{occ}}(\Delta t)$ in the current estimated body frame. Their forward, backward, left, and right extrema determine $\ell_i^{\mathrm{proj},t}$, from which $B_i^{\mathrm{proj},t}$ is constructed as the smallest current-frame rectangle enclosing both footprints. Because the projection depends on the estimated motion state and the response limits in $\eta_i$, the projected-motion layer expands according to both current motion and agent-specific actuation response.

\subsubsection{Stopping-Reservation Layer}

The stopping-reservation layer reserves the additional space required for an emergency stop from the projected state. The canonical stopping-distance model $d=v^2/(2a)$ assumes immediate constant deceleration and therefore does not account for the finite jerk and response latency represented by our actuation model. Consistent with the response limits in Eq.~\ref{eq:actuation_constraints}, we compute stopping distance from the terminal projected speed $\tilde{s}_i(\Delta t)$ and longitudinal acceleration $\tilde{\alpha}_i(\Delta t)$ obtained in Eq.~\ref{eq:projected_terminal_state}.

During the response latency $\tau_i$, the agent continues under $\tilde{\alpha}_i(\Delta t)$. Braking then drives the longitudinal acceleration toward $-a_i^{\max}$ at the maximum permitted jerk $j_i^{\max}$ and maintains maximum deceleration, if necessary, until the speed reaches zero. Let $T_i^{\mathrm{stop}}$ denote this first zero-speed time. The corresponding stopping distance is $d_i^{\mathrm{stop},t}=\int_{0}^{T_i^{\mathrm{stop}}} s_i^{\mathrm{stop}}(\xi)\,d\xi$, with $s_i^{\mathrm{stop}}(0)=\tilde{s}_i(\Delta t)$ and $\alpha_i^{\mathrm{stop}}(0)=\tilde{\alpha}_i(\Delta t)$.

As shown in Fig.~\ref{fig:dobb_stopping}, the terminal occupancy footprint is extended forward along the projected heading by $d_i^{\mathrm{stop},t}$ to define $B_i^{\mathrm{stop},t}$. The stopping-reservation layer $B_i^{\mathrm{res},t}$ is then constructed as the smallest current-frame rectangle enclosing $B_i^{\mathrm{proj},t}$ and $B_i^{\mathrm{stop},t}$. 

%% file: 5b_Observations.tex
\subsection{Host and Pairwise Geometric Features}
\label{subsec:pairwise-features}

\input{artifacts/fig_observation_features}
\input{artifacts/tab_face_gaps_interpretation}

We refer to agent $i$ as the host and agent $j$ as a locally observed neighbor. The host layer extents describe its own layered safety geometry, which we collect as
\begin{equation}
h_{i,\mathrm{DOBB}}^t
=
\ell_i^{\mathrm{occ},t}
\oplus
\ell_i^{\mathrm{proj},t}
\oplus
\ell_i^{\mathrm{res},t},
\label{eq:host_dobb_features}
\end{equation}
where $\oplus$ denotes vector concatenation.
Collision avoidance additionally requires pairwise features that describe how each neighbor interacts with this geometry. Since the host selects only its own action, we express these interactions relative to the full safety region it seeks to preserve, $B_i^{\mathrm{res},t}$. We compare this host reservation with each neighbor layer $B_j^{z,t}$, $z\in\mathcal Z$, which preserves the distinction between the neighbor's current occupancy, projected motion, and stopping reservation.A useful interaction descriptor must retain where the neighbor lies relative to the host, the clearance or penetration associated with that placement, whether the oriented regions actually intersect, and the severity of any encroachment. We represent these properties through directional face gaps $g_{ij}^{z,t}$, a signed separating-axis margin $c_{ij}^{z,t}$, and normalized area encroachment $\kappa_{ij}^{z,t}$. Their joint interpretation is summarized in Table~\ref{tab:geometric_interpretation}.

Directional face gaps are obtained by projecting the corners of $B_j^{z,t}$ onto the host-forward and host-left axes relative to $\hat p_i^t$, as illustrated in Fig.~\ref{fig:dobb_pairwise_projection}. Let $[x_z^-,x_z^+]$ and $[y_z^-,y_z^+]$ denote the resulting longitudinal and lateral intervals. The normalization factors are $\nu_x^z=\ell_{i,f}^{\mathrm{res},t}+\ell_{i,b}^{\mathrm{res},t}+x_z^+-x_z^-$ and $\nu_y^z=\ell_{i,l}^{\mathrm{res},t}+\ell_{i,r}^{\mathrm{res},t}+y_z^+-y_z^-$. The normalized gaps are
\begin{equation}
g_{ij}^{z,t}
=
\begin{bmatrix}
(x_z^- - \ell_{i,f}^{\mathrm{res},t})/\nu_x^z\\
(-\ell_{i,b}^{\mathrm{res},t}-x_z^+)/\nu_x^z\\
(y_z^- - \ell_{i,l}^{\mathrm{res},t})/\nu_y^z\\
(-\ell_{i,r}^{\mathrm{res},t}-y_z^+)/\nu_y^z
\end{bmatrix}.
\label{eq:directional_face_gaps}
\end{equation}
These gaps retain the host-relative direction of the interaction together with the normalized projected clearance or penetration from each host face, as illustrated in Fig.~\ref{fig:dobb_pairwise_observations}.

Actual intersection of the oriented rectangles is determined separately using the standard Separating Axis Theorem (SAT). Let $\mathcal A_{ij}^{z,t}$ contain the edge-normal axes of $B_i^{\mathrm{res},t}$ and $B_j^{z,t}$. If their projection intervals onto axis $n$ are
$[\underline{\rho}_i(n),\overline{\rho}_i(n)]$
and
$[\underline{\rho}_j(n),\overline{\rho}_j(n)]$, respectively, the signed SAT margin is
\begin{equation}
c_{ij}^{z,t}
=
\max_{n\in\mathcal A_{ij}^{z,t}}
\max
\left\{
\underline{\rho}_j(n)-\overline{\rho}_i(n),
\underline{\rho}_i(n)-\overline{\rho}_j(n)
\right\}.
\label{eq:sat_margin}
\end{equation}
Positive $c_{ij}^{z,t}$ indicates separation, zero indicates boundary contact, and negative $c_{ij}^{z,t}$ indicates intersection. When the regions intersect, $-c_{ij}^{z,t}$ gives the minimum separating translation along the tested axes.

The SAT margin measures separation or penetration along the critical axis, but does not indicate how much of the neighbor layer encroaches on the host reservation. We therefore define the normalized area encroachment
\begin{equation}
\kappa_{ij}^{z,t}
=
\frac{
\operatorname{area}
\left(
B_i^{\mathrm{res},t}\cap B_j^{z,t}
\right)
}{
\operatorname{area}
\left(
B_j^{z,t}
\right)
}.
\label{eq:normalized_encroachment}
\end{equation}
The value satisfies $0\leq\kappa_{ij}^{z,t}\leq1$, with larger values indicating that a greater fraction of the neighbor layer lies inside the host reservation. Normalization by the neighbor-layer area keeps the measure comparable as uncertainty, physical size, and response limits alter DOBB geometry. The measure is intentionally asymmetric because the interaction is evaluated from the host perspective. Fig.~\ref{fig:dobb_pairwise_observations} and Table~\ref{tab:geometric_interpretation} summarize how the three quantities jointly describe placement, separation, intersection, and encroachment severity.

The complete DOBB pairwise descriptor is defined by
\begin{equation}
\begin{aligned}
q_{ij,\mathrm{DOBB}}^t
={}&
\left(\bigoplus_{z\in\mathcal Z}\kappa_{ij}^{z,t}\right)
\oplus
\left(\bigoplus_{z\in\mathcal Z}c_{ij}^{z,t}\right)
\oplus
\left(\bigoplus_{z\in\mathcal Z}g_{ij}^{z,t}\right),
\end{aligned}
\label{eq:pairwise_dobb_features}
\end{equation}
and along with, $h_{i,\mathrm{DOBB}}^t$ and $q_{ij,\mathrm{DOBB}}^t$ expose the host safety geometry and the direction and severity of its interaction with each neighbor.

%% file: artifacts/fig_observation_features.tex
\begin{figure}[t]
    \centering

    \begin{subfigure}[t]{0.485\columnwidth}
        \centering
        \includegraphics[width=\linewidth]
        {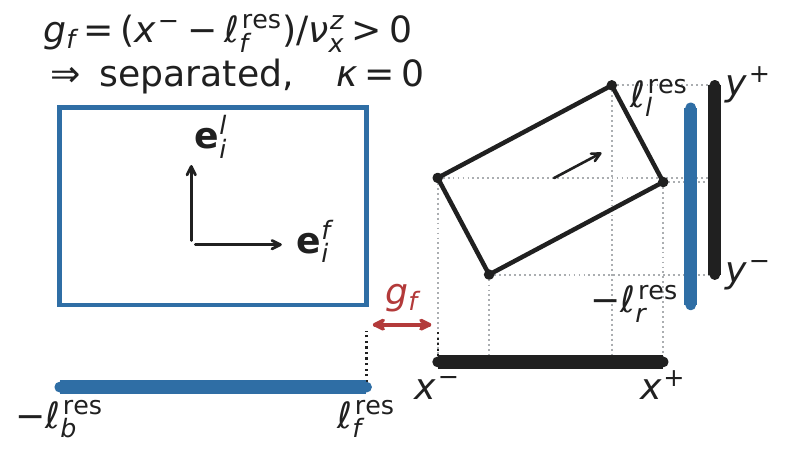}
        \caption{}
        \label{fig:dobb_pairwise_projection}
    \end{subfigure}
    \hfill
    \begin{subfigure}[t]{0.485\columnwidth}
        \centering
        \includegraphics[width=\linewidth]
        {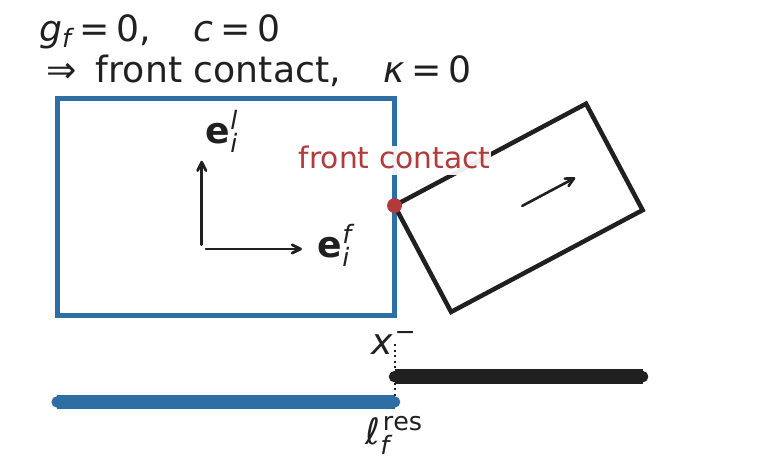}
        \caption{}
        \label{fig:dobb_pairwise_contact}
    \end{subfigure}

    \begin{subfigure}[t]{0.485\columnwidth}
        \centering
        \includegraphics[width=\linewidth]
        {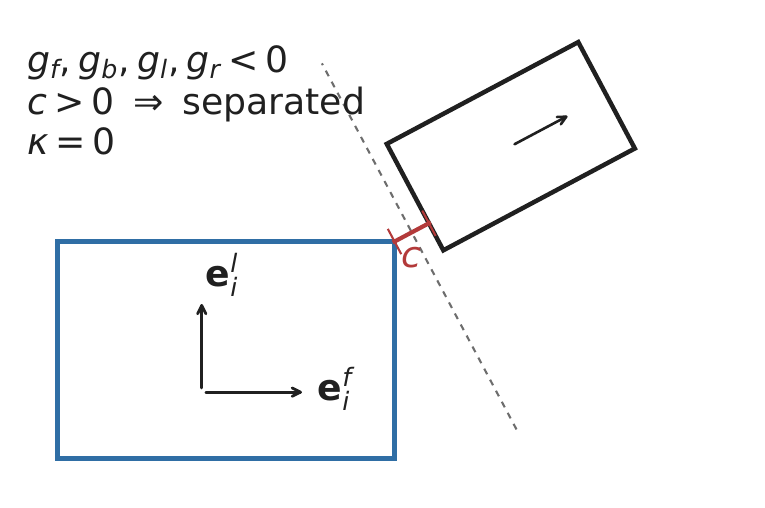}
        \caption{}
        \label{fig:dobb_pairwise_sat_separation}
    \end{subfigure}
    \hfill
    \begin{subfigure}[t]{0.485\columnwidth}
        \centering
        \includegraphics[width=\linewidth]
        {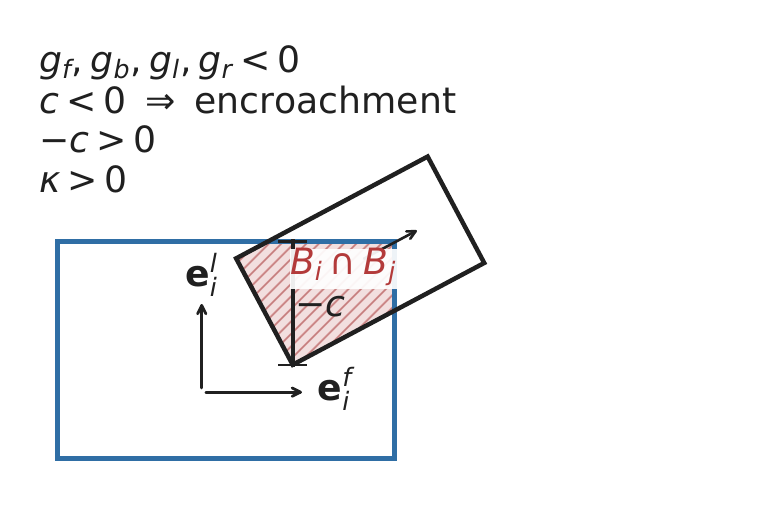}
        \caption{}
        \label{fig:dobb_pairwise_encroachment}
    \end{subfigure}

    \caption{DOBB pairwise geometric observations: (a) projected face-gap separation; (b) front contact; (c) SAT-certified separation; and (d) SAT-confirmed encroachment.}
    \label{fig:dobb_pairwise_observations}
\end{figure}

%% file: artifacts/tab_face_gaps_interpretation.tex

\begin{table}[t]
\centering
\caption{Joint geometric interpretation of directional face gaps $g_{ij}^{z,t}$, SAT margin $c_{ij}^{z,t}$, and normalized encroachment $\kappa_{ij}^{z,t}$.}
\label{tab:geometric_interpretation}

\footnotesize
\renewcommand{\arraystretch}{1.08}
\setlength{\tabcolsep}{4pt}

\begin{tabularx}{\columnwidth}{
    @{}
    >{\raggedright\arraybackslash}p{0.17\columnwidth}
    >{\raggedright\arraybackslash}X
    @{}
}
\toprule
\textbf{Condition} & \textbf{Geometric interpretation} \\
\midrule

$g_k > 0$
&
$B_j$ lies entirely beyond host face $k$, so $B_i$ and $B_j$ are
disjoint ($c>0,\ \kappa=0$). $g_k$ gives normalized projected
clearance from face $k$; multiple positive components indicate
compound placement, e.g., front-left.
\\

\addlinespace[2pt]

$g_k = 0$
&
A boundary of the $B_j$ projection coincides with the supporting
line of host face $k$. If $c>0$, another SAT axis separates
$B_i$ and $B_j$; if $c=0$, they are in boundary contact involving
face $k$. In either case, $\kappa=0$.
\\

\addlinespace[2pt]

$\mathbf{g} < 0$
&
The projections of $B_i$ and $B_j$ overlap along both host axes.
$c>0$ indicates separation along a neighbor-aligned SAT axis,
$c=0$ boundary contact, and $c<0$ strict intersection. For $c<0$,
$-c$ gives the minimum separating translation and $\kappa>0$
quantifies normalized encroachment.
\\

\bottomrule
\end{tabularx}

\vspace{2pt}
\parbox{\columnwidth}{\scriptsize
$B_i\equiv B_i^{\mathrm{res},t}$,
$B_j\equiv B_j^{z,t}$ for $z\in\mathcal Z$, and
$k\in\{f,b,l,r\}$. We write $\mathbf{g}<0$ when all four face gaps
are negative. Within each opposite-face pair $(k,\bar{k})$, the
component closer to zero identifies the nearer host face and smaller
host-axis separating translation.
}
\vspace{-12pt}
\end{table}

%% file: 5c_Reward.tex
\subsection{Encroachment Penalty}
\label{subsec:reward}
The observation features retain continuous information about interaction direction, clearance, penetration, and encroachment. The reward assigns physical significance to these interactions by penalizing encroachment according to the DOBB layer involved. Encroachment with $B_j^{\mathrm{res},t}$ indicates that the two agents can no longer simultaneously preserve their full braking space. At this stage, the interaction warrants caution yet remains recoverable. Encroachment with $B_j^{\mathrm{proj},t}$ marks a more urgent conflict, as the overlap now involves motion already committed for execution and affords less opportunity to recover separation before that motion unfolds. Encroachment with $B_j^{\mathrm{occ},t}$ represents the most immediate pre-collision condition. The host stopping reservation now intersects a region in which the neighbor may physically be present at the current step, making continued approach directly collision-relevant. Because the DOBB layers are nested, we assign the penalty according
to the innermost neighbor layer with positive-area encroachment:
\begin{equation}
r_{ij,\mathrm{enc}}^t =
\begin{cases}
-\lambda_{\mathrm{occ}},
& \kappa_{ij}^{\mathrm{occ},t}>0, \\[2pt]
-\lambda_{\mathrm{proj}},
& \kappa_{ij}^{\mathrm{occ},t}=0
  \ \land\ 
  \kappa_{ij}^{\mathrm{proj},t}>0, \\[2pt]
-\lambda_{\mathrm{res}},
& \kappa_{ij}^{\mathrm{proj},t}=0
  \ \land\ 
  \kappa_{ij}^{\mathrm{res},t}>0, \\[2pt]
0,
& \text{otherwise}.
\end{cases}
\label{eq:encroachment_penalty}
\end{equation}

The nesting
$B_j^{\mathrm{occ},t}\subseteq B_j^{\mathrm{proj},t}\subseteq B_j^{\mathrm{res},t}$
ensures that Eq.~\ref{eq:encroachment_penalty} selects the most
severe active encroachment. For multiple neighbors, the host receives the most severe pairwise penalty, $r_{i,\mathrm{enc}}^t=\min(0,\min_{j\in\mathcal N_i^t}r_{ij,\mathrm{enc}}^t)$. This prevents the penalty from increasing simply with neighbor count and keeps the reward tied to the most critical interaction. The numerical values of $\lambda_{\mathrm{occ}},\lambda_{\mathrm{proj}},\lambda_{\mathrm{res}}$ are reported with the remaining policy parameters in Sec.~\ref{subsec:exp-setup_pol-reward}.

%% file: 6_Experimental_Setup.tex
\section{Experimental Setup}
\label{sec:experimental-setup}
\subsection{Policy \& Reward Instantiation}
\label{subsec:exp-setup_pol-reward}

To isolate the contribution of DOBB from changes to the learning architecture, we instantiate it within the open-source GA3C-CADRL framework of \cite{everett2021collision}. We retain its neighbor LSTM, shared actor-critic network, and 11-action speed-and-heading space. DOBB modifies the policy observation and nonterminal geometric reward while preserving the underlying network and action space.

The base observation consists of a host descriptor $h_{i,\mathrm{base}}^t$ and one pairwise descriptor $q_{ij,\mathrm{base}}^t$ for each observed neighbor,
\begin{equation}
h_{i,\mathrm{base}}^t=
[d_{g,i}^t,\psi_{g,i}^t,v_i^{\mathrm{pref}},r_i]^\top,
\label{eq:base_host_observation}
\end{equation}
\begin{equation}
q_{ij,\mathrm{base}}^t=
[p_{ij}^{f,t},p_{ij}^{l,t},v_{ij}^{f,t},v_{ij}^{l,t},r_j,r_{ij},d_{ij}^{\mathrm{phys},t}]^\top.
\label{eq:base_pairwise_observation}
\end{equation}

Here, $d_{g,i}^t$ is the host-goal distance and $\psi_{g,i}^t$ is the angular difference between the host heading and goal bearing. The pairwise descriptor contains host-frame relative position $(p_{ij}^{f,t},p_{ij}^{l,t})$, relative velocity $(v_{ij}^{f,t},v_{ij}^{l,t})$, neighbor physical radius $r_j$, combined physical radius $r_{ij}=r_i+r_j$, and physical-footprint clearance $d_{ij}^{\mathrm{phys},t}=\|\hat p_j^t-\hat p_i^t\|_2-r_{ij}$.

DOBB augments these descriptors with the host and pairwise geometric features defined in Sec.~\ref{sec:methodology},
\begin{equation}
h_i^t=h_{i,\mathrm{base}}^t\oplus h_{i,\mathrm{DOBB}}^t,
\qquad
q_{ij}^t=q_{ij,\mathrm{base}}^t\oplus q_{ij,\mathrm{DOBB}}^t.
\label{eq:dobb_augmented_observation}
\end{equation}

The neighbor vectors remain ordered from farther to nearer agents and are processed by the same LSTM and actor-critic network.

GA3C-CADRL assigns $+1$ upon goal arrival and $-0.25$ upon physical collision. Its nonterminal reward penalizes physical proximity according to
\begin{equation}
r_{i,\mathrm{prox}}^t=
\begin{cases}
-0.1-\dfrac{d_{i,\min}^t}{2}, & 0<d_{i,\min}^t\leq0.2,\\
0, & \text{otherwise},
\end{cases}
\label{eq:baseline_proximity_reward}
\end{equation}
where $d_{i,\min}^t$ is the minimum physical-footprint clearance to any neighbor. This term imposes the same $0.2$ m proximity threshold for every interaction, independent of state uncertainty or heterogeneous response. DOBB retains the goal and collision rewards and replaces this nonterminal term with the encroachment reward $r_{i,\mathrm{enc}}^t$ defined in Eq.~\ref{eq:encroachment_penalty}, using $(\lambda_{\mathrm{occ}},\lambda_{\mathrm{proj}},\lambda_{\mathrm{res}})=(0.12,0.04,0.01)$.

\subsection{Training and Operational Design Domain}

We retain the training procedure of \cite{everett2021collision}. The policy is initialized through supervised learning from CADRL state-action-value tuples and subsequently refined through reinforcement-learning phases with up to four and ten agents. Agent count, initial and goal positions, physical radius, preferred speed, and the policies assigned to nonlearning agents remain randomized during training.

The full training environment extends the original framework in three respects. First, the original GA3C-CADRL assumption that selected speed and heading commands take effect immediately is replaced by the actuation-constrained model of Sec.~\ref{subsec:actuation-constrained-motion}, where commands are realized subject to agent-specific acceleration, jerk, yaw-rate, and yaw-acceleration limits. Second, zero-mean Gaussian noise is added to position and speed measurements, and state estimation is performed using an extended Kalman filter (EKF). EKFs remain widely used for state estimation in UAV autopilots \cite{10611108}; DOBB itself requires only an estimated state and an associated uncertainty description. Third, the heterogeneous parameters in $\eta_i$ are sampled independently for each agent.

Table~\ref{tab:training_parameters} defines the Env4 training distribution through the allowable ranges of $\eta_i$ and the position- and speed-noise regimes. We denote this full configuration as Env4 in Sec.~\ref{sec:evaluation}. Conditions outside these ranges are reserved for the out-of-distribution (OOD) evaluations.

\input{artifacts/tab_training_distributions}

%% file: artifacts/tab_training_distributions.tex

\begin{table}[t]
\centering
\caption{Agent-parameter ranges and measurement-noise regimes used for $\mathrm{Env}_4$ training.}
\label{tab:training_parameters}

\footnotesize
\setlength{\tabcolsep}{5pt}
\renewcommand{\arraystretch}{1.12}

\begin{tabular}{@{}lc@{}}
\toprule
\textbf{Parameter} & \textbf{Training range} \\
\midrule

Preferred speed $v_i^{\mathrm{pref}}$
& $[0.5,\,2.0]~\mathrm{m/s}$ \\

Physical radius $r_i$
& $[0.2,\,0.8]~\mathrm{m}$ \\

Maximum acceleration $a_i^{\max}$
& $[2.5,\,3.5]~\mathrm{m/s^2}$ \\

Maximum jerk $j_i^{\max}$
& $[6.0,\,10.0]~\mathrm{m/s^3}$ \\

Maximum yaw rate $\omega_i^{\max}$
& $[0.9,\,1.5]~\mathrm{rad/s}$ \\

Maximum yaw acceleration $\dot{\omega}_i^{\max}$
& $[3.0,\,5.0]~\mathrm{rad/s^2}$ \\

Response latency $\tau_i$
& $[0.03,\,0.08]~\mathrm{s}$ \\

\midrule

\multirow{3}{*}{%
    \shortstack[c]{Noise regimes\\$\mathrm{Probability}:(\sigma_p~[\mathrm{m}],\,\sigma_s~[\mathrm{m/s}])$}
}
& $0.30:(0,\,0)$ \\

& $0.50:([0.05,\,0.13],\,[0.02,\,0.05])$ \\

& $0.20:([0.14,\,0.20],\,[0.05,\,0.07])$ \\

\bottomrule
\end{tabular}
\vspace{-12pt}
\end{table}

%% file: 7_Evaluation.tex
\vspace{-8pt}
\section{Evaluation}
\label{sec:evaluation}
We evaluate how the representation of state-estimation uncertainty and agent heterogeneity affects collision-avoidance performance under both in-distribution (ID) and out-of-distribution (OOD) conditions. The evaluation addresses two research questions.

\textbf{RQ1:} How does collision-avoidance performance change across progressively higher-fidelity training environments and when uncertainty and heterogeneity are represented through experience-based learning, explicit-runtime information, or DOBB geometry?

\textbf{RQ2:} How well do these representations generalize as state-estimation uncertainty moves beyond the training distribution?

To capture training variability, each policy is trained independently three times, with one checkpoint per run selected on a disjoint validation set. The selected policies are evaluated on 500 ten-agent test scenarios. We report mutually exclusive scenario-level rates for collision only, timeout only, and joint collision-and-timeout outcomes, with total failure defined as their union. Results are summarized by the mean and sample standard deviation across the three runs.

\input{7a_RQ1}

\input{7b_RQ2}

%% file: 7a_RQ1.tex
\subsection{RQ1: Training Ablations \& Representation of Uncertainty and Heterogeneity}
\label{subsec:rq1}
\input{artifacts/tab_rq1_training_environments}
\input{artifacts/fig_eval_exp2_v2}
This section disentangles two effects that are otherwise easily confounded. We first examine the effect of training-domain exposure. As summarized in Table~\ref{tab:rq1_training_environments}, each policy $A_i$ is trained only in its corresponding $\mathrm{Env}_i$, where the nested environments progressively introduce actuation-constrained dynamics, state-estimation uncertainty, and heterogeneous agent capabilities while retaining the same base observation and proximity reward. Evaluating each $A_i$ across all four environments then reveals how its training-domain exposure affects robustness when the corresponding conditions are present, absent, or shifted at inference. The second comparison fixes training to $\mathrm{Env}_4$ for all formulations and varies how state-estimation uncertainty and heterogeneous actuation are incorporated into the policy formulation under the experience-based, explicit-runtime, and DOBB approaches introduced in Secs.~\ref{sec:introduction}--\ref{sec:related-work}. $A_4$ uses only the base observation and serves as the experience-based reference, whose response to uncertainty and actuation limits is learned from training interactions. $A_4^{\mathrm{belief}}$ provides the explicit-runtime comparison by augmenting the host and pairwise observations with $b_i^t$ and $b_{ij}^t$, respectively. For compactness, let
$\eta_i^{\mathrm{act}}\equiv
(a_i^{\max},j_i^{\max},\omega_i^{\max},\dot{\omega}_i^{\max},\tau_i)$
denote the actuation-response subset of $\eta_i$ defined in Sec.~\ref{sec:problem-setup}. Then
\begin{equation}
b_i^t =
\left[
(\hat{s}_i^t,\hat{\alpha}_i^t,\hat{\omega}_i^t),\,
(\delta_{i,f}^{t},\delta_{i,l}^{t}),\,
\eta_i^{\mathrm{act}}
\right]^\top 
\label{eq:belief_host_features}
\end{equation}
and
\begin{equation}
b_{ij}^t =
\left[
(\delta_{ij,f}^{t},\delta_{ij,l}^{t}),\,
(a_{ij,f}^t,a_{ij,l}^t),\,
(\hat{s}_j^t,\hat{\alpha}_j^t,\hat{\omega}_j^t),\,
v_j^{\mathrm{pref}},\,
\eta_j^{\mathrm{act}}
\right]^\top .
\label{eq:belief_pairwise_features}
\end{equation}

Here, $\hat{s}^t$, $\hat{\alpha}^t$, and $\hat{\omega}^t$ denote estimated speed, longitudinal acceleration, and yaw rate; $\delta_{i,f}^{t}$ and $\delta_{i,l}^{t}$ are the host uncertainty bounds defined in Sec.~\ref{subsec:dobb}, while $\delta_{ij,f}^{t}$ and $\delta_{ij,l}^{t}$ are the corresponding host-frame relative-position bounds computed from the summed host--neighbor covariance; lastly, $a_{ij,f}^t$ and $a_{ij,l}^t$ denote neighbor acceleration relative to the host along the same axes. These features expose uncertainty and agent response characteristics directly to $A_4^{\mathrm{belief}}$, while their implications for required physical separation remain learned by the policy.  $A_4^{\mathrm{DOBB\text{-}Obs}}$ maps these factors into the DOBB geometry but retains the baseline proximity reward to isolate the contribution of the geometric observation. $A_4^{\mathrm{DOBB}}$ additionally uses the proposed encroachment reward and represents the complete DOBB formulation. The network architecture and action space are held fixed across the formulations.

\input{artifacts/tab_rq1_results_compressed}

Table~\ref{tab:rq1_results_matrix} identifies actuation mismatch as the dominant source of degradation. \(A_1\), trained under instantaneous motion, increases from 1.8\% failure in \(\mathrm{Env}_1\) to more than 80\% in \(\mathrm{Env}_2\)--\(\mathrm{Env}_4\). Conversely, \(A_2\), \(A_3\), and \(A_4^{\mathrm{belief}}\), all trained with dynamics of Sec \ref{subsec:actuation-constrained-motion}, deteriorate sharply when evaluated in \(\mathrm{Env}_1\), reaching up to 54\% failure. Since \(\mathrm{Env}_1\) and \(\mathrm{Env}_2\) differ only in the actuation model, this cross-domain degradation isolates actuation mismatch as the principal source of distribution shift. Once constrained dynamics are represented during training, further increases in training fidelity yield comparatively small gains: \(A_4\) improves over \(A_2\) by only 1.7--1.9 percentage points across \(\mathrm{Env}_2\)--\(\mathrm{Env}_4\).. Thus, exposure to noisy estimation and heterogeneous capabilities improves robustness, but does not by itself resolve how these factors should alter collision-avoidance behavior.

Holding the training environment fixed at $\mathrm{Env}_4$ reveals a stronger effect from representation. $A_4^{\mathrm{belief}}$ reduces total failure from approximately 10--11\% for $A_4$ to 4--6\% across $\mathrm{Env}_2$--$\mathrm{Env}_4$, which suggests that explicit runtime access to uncertainty and heterogeneous capabilities is substantially more effective than training exposure alone. However, failures remain predominantly collision-driven: collision outcomes account for more than 71-86\% of failures for both policies in $\mathrm{Env}_3$ and $\mathrm{Env}_4$. Mapping these factors into DOBB geometry while retaining the baseline reward, $A_4^{\mathrm{DOBB\text{-}Obs}}$, further reduces total failure to 3.6--4.5\% across $\mathrm{Env}_2$--$\mathrm{Env}_4$. This improvement shows that the geometric abstraction alone provides additional benefit beyond explicit runtime inputs. However, $A_4^{\mathrm{DOBB\text{-}Obs}}$ remains susceptible to the actuation-domain shift in $\mathrm{Env}_1$, where total failure rises to 13.8\%, and its failures remain overwhelmingly collision-driven. 

The encroachment reward changes this failure profile. In the complete $A_4^{\mathrm{DOBB}}$ formulation, it achieves the lowest mean failure, $ \bar F = 2.6\% $, with a collision-only failure of $\bar C = 2.6\% $ averaged over all four environments. In $\mathrm{Env}_4$, this corresponds to collision-reductions of approximately 79\%, 61\%, and 49\% relative to $A_4$, $A_4^{\mathrm{belief}}$, and $A_4^{\mathrm{DOBB\text{-}Obs}}$, respectively. The tradeoff appears as a modest and arguably negligible increase in total failure relative to $A_4^{\mathrm{DOBB\text{-}Obs}}$, but compromised as timeouts rather than collisions. The full formulation also avoids the sharp actuation-shift degradation exhibited by $A_2$, $A_3$, $A_4^{\mathrm{belief}}$, and $A_4^{\mathrm{DOBB\text{-}Obs}}$. In $\mathrm{Env}_1$, $A_4^{\mathrm{DOBB}}$ achieves 2.33\% total failure, compared with 13.8\% for $A_4^{\mathrm{DOBB\text{-}Obs}}$, and approaches the 1.8\% failure of $A_1$ in its native environment. A plausible explanation is that DOBB encodes changes in uncertainty and actuation response through the geometry itself. The policy therefore learns avoidance from separation and encroachment relationships whose interpretation remains consistent across actuation regimes.

%% file: artifacts/tab_rq1_training_environments.tex


\begin{table}[t]
\centering
\caption{Training environments used for the RQ1 training-fidelity ablation.}
\label{tab:rq1_training_environments}

\footnotesize
\setlength{\tabcolsep}{5pt}
\renewcommand{\arraystretch}{1.12}

\begin{tabular}{@{}lccc@{}}
\toprule

\textbf{Environment}
&
\shortstack{\textbf{Constrained}\\\textbf{dynamics}}

&
\shortstack{\textbf{Heterogeneous}\\\textbf{capabilities}}
&
\shortstack{\textbf{Noisy sensing}\\\textbf{and EKF}}

\\

\midrule

$\mathrm{Env}_1$
& --
& --
& --
\\

$\mathrm{Env}_2$
& $\checkmark$
& --
& --
\\

$\mathrm{Env}_3$
& $\checkmark$
& $\checkmark$
& --
\\

$\mathrm{Env}_4$
& $\checkmark$
& $\checkmark$
& $\checkmark$
\\

\bottomrule
\end{tabular}
\vspace{-8pt}
\end{table}


\begin{table}[t]
\centering
\caption{Policy configurations used to compare uncertainty and heterogeneity representations in RQ1. Reward denotes the nonterminal geometric term.}
\label{tab:rq1_policy_instantiations}

\footnotesize
\setlength{\tabcolsep}{3pt}
\renewcommand{\arraystretch}{1.12}

\begin{tabularx}{
    \columnwidth
}{
    @{}
    l
    c
    >{\raggedright\arraybackslash}X
    >{\raggedright\arraybackslash}X
    @{}
}
\toprule

\textbf{Policy}
&
\shortstack{\textbf{Training}\\\textbf{environment}}
&
\textbf{Observations}
&
\textbf{Reward}
\\
\midrule

\shortstack[l]{
    $A_m$\\[-1pt]
    $(m=1,\ldots,4)$
}
&
$\mathrm{Env}_m$
&
Eqs.~\ref{eq:base_host_observation}--\ref{eq:base_pairwise_observation}
&
Eq.~\ref{eq:baseline_proximity_reward}
\\[2pt]

$A_4^{\mathrm{belief}}$
&
$\mathrm{Env}_4$
&
Eqs.~\ref{eq:base_host_observation}--\ref{eq:base_pairwise_observation} \&
Eqs.~\ref{eq:belief_host_features}--\ref{eq:belief_pairwise_features}
&
Eq.~\ref{eq:baseline_proximity_reward}
\\[2pt]

$A_4^{\mathrm{DOBB\text{-}Obs}}$
&
$\mathrm{Env}_4$
&
Eq.~\ref{eq:dobb_augmented_observation}
&
Eq.~\ref{eq:baseline_proximity_reward}
\\[2pt]

$A_4^{\mathrm{DOBB}}$
&
$\mathrm{Env}_4$
&
Eq.~\ref{eq:dobb_augmented_observation}
&
Eq.~\ref{eq:encroachment_penalty}
\\

\bottomrule
\end{tabularx}
\vspace{-12pt}
\end{table}

%% file: artifacts/fig_eval_exp2_v2.tex
\begin{figure*}[t]
    \centering
    \subfloat[Total failure]{
        \includegraphics[width=0.32\textwidth]
        {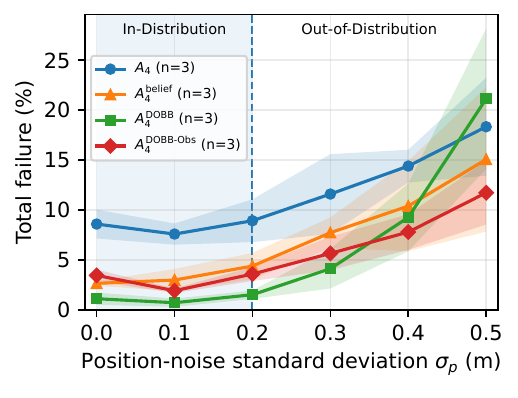}
        \label{fig:rq2_total_failure}
    }
    \subfloat[Failure-mode decomposition.]{
        \includegraphics[width=0.31\textwidth]
        {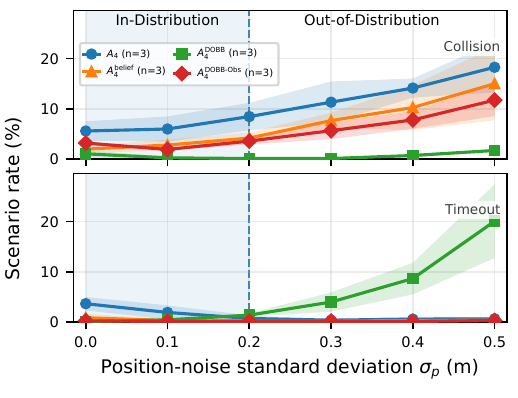}
        \label{fig:rq2_failure_modes}
    }
    \subfloat[Example trajectories.]{
        \includegraphics[width=0.32\textwidth]
        {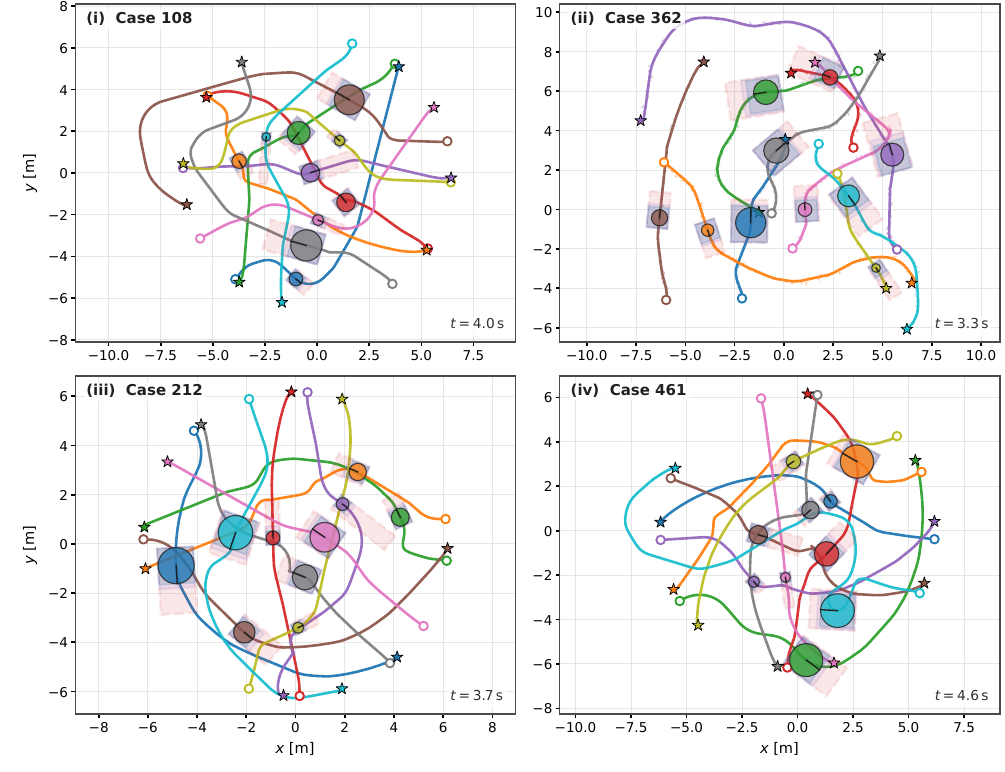}
        \label{fig:rq2_third_figure}
    }

    \caption{RQ2 generalization under increasing position-noise standard deviation $\sigma_p$. The blue background denotes the in-distribution training-noise range up to $\sigma_p=0.20~\mathrm{m}$; $\sigma_p>0.20~\mathrm{m}$ is out of distribution.}
    \label{fig:rq2_uncertainty_generalization}
    \vspace{-12pt}
\end{figure*}

%% file: artifacts/tab_rq1_results_compressed.tex
\newcommand{\rqbest}[1]{\textbf{#1}}

\newcommand{\rqtotal}[2]{\( #1 \pm #2 \)}

\newcommand{\rqbesttotal}[2]{%
    {\bfseries\boldmath \( #1 \pm #2 \)}%
}

\newcommand{\rqcell}[4]{%
    \shortstack[c]{%
        {\scriptsize #4}\\[-1pt]
        {\footnotesize #1/#2/#3}%
    }%
}

\newcommand{\rqavgcell}[2]{%
    \raisebox{1.3pt}{%
        \shortstack[c]{%
            {\scriptsize \(#1\)}\\[-1pt]
            {\footnotesize #2}%
        }%
    }%
}

\newcommand{\rqbestavgcell}[2]{%
    \raisebox{0.5pt}{%
        \shortstack[c]{%
            {\scriptsize\bfseries\boldmath \(#1\)}\\[-1pt]
            {\footnotesize\bfseries #2}%
        }%
    }%
}

\begin{table}[t]
\centering
\caption{RQ1 failure rates by policy and evaluation environment.
Each cell reports $F\pm\sigma_F$ and $C/T/J$, in percent.}
\label{tab:rq1_results_matrix}

\footnotesize
\renewcommand{\arraystretch}{2.00}
\setlength{\tabcolsep}{1.0pt}

\begin{tabular}{@{}lccccc@{}}
\toprule
\textbf{Policy}
&
\(\mathrm{Env}_1\)
&
\(\mathrm{Env}_2\)
&
\(\mathrm{Env}_3\)
&
\(\mathrm{Env}_4\)
&
\shortstack[c]{%
    \(\bar F\)\\[-1pt]
    \(\bar C\)%
}
\\
\midrule

\(A_1\) \cite{everett2021collision}
&
\rqcell{\rqbest{1.0}}{0.8}{0.0}
       {\rqbesttotal{1.80}{0.53}}
&
\rqcell{69.3}{3.1}{8.5}
       {\rqtotal{80.80}{3.34}}
&
\rqcell{69.7}{3.8}{9.6}
       {\rqtotal{83.07}{4.50}}
&
\rqcell{71.0}{3.8}{10.4}
       {\rqtotal{85.20}{4.26}}
&
\rqavgcell{62.7}{52.8}
\\

\(A_2\)
&
\rqcell{2.5}{20.6}{0.8}
       {\rqtotal{23.93}{17.90}}
&
\rqcell{6.3}{5.0}{0.7}
       {\rqtotal{12.00}{0.35}}
&
\rqcell{7.3}{4.6}{1.3}
       {\rqtotal{13.20}{0.72}}
&
\rqcell{9.3}{2.1}{0.9}
       {\rqtotal{12.33}{1.96}}
&
\rqavgcell{15.4}{6.4}
\\

\(A_3\)
&
\rqcell{1.1}{51.8}{1.2}
       {\rqtotal{54.07}{42.27}}
&
\rqcell{5.3}{5.6}{0.4}
       {\rqtotal{11.33}{0.92}}
&
\rqcell{6.5}{5.5}{1.2}
       {\rqtotal{13.13}{1.63}}
&
\rqcell{8.1}{4.3}{0.7}
       {\rqtotal{13.13}{0.50}}
&
\rqavgcell{22.9}{5.3}
\\

\(A_4\)
&
\rqcell{3.6}{5.2}{0.7}
       {\rqtotal{9.53}{1.47}}
&
\rqcell{5.3}{4.2}{0.8}
       {\rqtotal{10.33}{2.25}}
&
\rqcell{7.4}{3.3}{0.7}
       {\rqtotal{11.40}{1.71}}
&
\rqcell{7.9}{2.0}{0.6}
       {\rqtotal{10.47}{0.95}}
&
\rqavgcell{10.4}{6.1}
\\

\midrule

\(A_4^{\mathrm{belief}}\)
&
\rqcell{10.5}{10.3}{1.7}
       {\rqtotal{22.53}{15.21}}
&
\rqcell{2.2}{1.5}{0.2}
       {\rqtotal{3.93}{0.61}}
&
\rqcell{4.7}{1.2}{0.3}
       {\rqtotal{6.13}{1.85}}
&
\rqcell{4.5}{0.7}{0.1}
       {\rqtotal{5.33}{0.99}}
&
\rqavgcell{9.5}{5.5}
\\

{\scriptsize \(A_4^{\mathrm{DOBB\text{-}Obs}}\)}
&
\rqcell{12.8}{0.2}{0.8}
       {\rqtotal{13.80}{1.91}}
&
\rqcell{3.3}{0.3}{0.1}
       {\rqtotal{3.60}{0.53}}
&
\rqcell{3.9}{0.3}{0.3}
       {\rqtotal{4.47}{1.03}}
&
\rqcell{3.2}{0.2}{0.3}
       {\rqbesttotal{3.67}{1.14}}
&
\rqavgcell{6.4}{5.8}
\\

\(A_4^{\mathrm{DOBB}}\)
&
\rqcell{2.3}{0.1}{0.0}
       {\rqtotal{2.33}{0.95}}
&
\rqcell{\rqbest{1.7}}{0.0}{0.0}
       {\rqbesttotal{1.73}{0.64}}
&
\rqcell{\rqbest{2.0}}{0.0}{0.0}
       {\rqbesttotal{2.00}{0.20}}
&
\rqcell{\rqbest{1.7}}{2.7}{0.1}
       {\rqtotal{4.47}{0.64}}
&
\rqbestavgcell{2.6}{1.9}
\\

\bottomrule
\end{tabular}

\vspace{2pt}

\parbox{\columnwidth}{%
\scriptsize
\raggedright
\(F\): total failure; \(C\), \(T\), and \(J\): collision-only,
timeout-only, and joint failure; \(\bar F\) and \(\bar C\): means across
evaluation environments. Bold indicates the lowest \(F\) and \(C\)
within each environment and the lowest corresponding averages;
tied minima are both bolded.
}

\vspace{-12pt}
\end{table}

%% file: 7b_RQ2.tex
\subsection{RQ2: Generalization Beyond Training Distribution}
\label{subsec:rq2}

To evaluate generalization beyond the training distribution, $A_4$, $A_4^{\mathrm{belief}}$, $A_4^{\mathrm{DOBB\text{-}Obs}}$, and $A_4^{\mathrm{DOBB}}$ are evaluated on the same 500 ten-agent $\mathrm{Env}_4$ scenarios while progressively increasing the position-noise standard deviation $\sigma_p$ beyond the ID training range. All other evaluation conditions are held fixed.

Within the ID range, Fig.~\ref{fig:rq2_total_failure} shows only modest sensitivity to increasing position noise, and the relative performance observed in RQ1 is preserved as $A_4^{\mathrm{belief}}$ consistently outperforms $A_4$ and $A_4^{\mathrm{DOBB}}$ remains below both. Beyond ID, $A_4$ and $A_4^{\mathrm{belief}}$ exhibit collision-dominated degradation, with collision-based failure reaching 18.27\% and 15.00\%, respectively, at $\sigma_p=0.5$~m. The DOBB observation alone is more robust to this shift. $A_4^{\mathrm{DOBB\text{-}Obs}}$ maintains lower total failure than both baselines at every tested noise level beyond ID; however, its failure profile remains collision-dominated and reaches  12\% at $\sigma_p=0.5$~m. The complete DOBB formulation changes this failure profile, where $A_4^{\mathrm{DOBB}}$ maintains collision-based failure below 2\% across all tested noise levels. Under severe uncertainty, its degradation occurs almost entirely through timeout behavior. This extends the RQ1 finding to a second form of distribution shift. DOBB avoids collision-driven degradation under both actuation mismatch and state-estimation uncertainty beyond the training distribution.

%% file: 8_Conclusion.tex
\vspace{-8pt}
\section{Conclusion}
\label{sec:conclusion}
Learning-based collision avoidance has primarily addressed state uncertainty and agent heterogeneity through training-domain variation or explicit runtime inputs. DOBB provides a complementary geometric representation that maps the current state estimate, uncertainty, and heterogeneous agent response into a dynamic, direction-dependent safety margin recomputed at every decision step. Furthermore, we derive geometric interaction features that characterize the relative configuration and severity of interactions between agents' DOBBs, along with a reward that preserves the physical interpretation of these interactions.

Across the evaluated training environments and OOD conditions (i.e. actuation-shifts \& high position uncertainty), DOBB maintains the lowest collision incidence, with collision-based failure at or below 2.3\% in every tested condition. We attribute this to DOBB's abstraction of uncertainty and agent response through safety geometry, which preserves consistent physical interpretations across different operating conditions. At sufficiently high uncertainty, however, the resulting safety regions consume more of the available free space and increasingly restrict admissible motion, which ultimitely shifts residual failure toward timeouts. This exposes a safety–progress tradeoff under severe uncertainty, with DOBB favoring safety until conditions return nominal. Code open-sourced at: \href{https://anonymous.4open.science/r/DOBB-Collision-Avoidance-8D16/README.md}{https://anonymous.4open.science/r/DOBB-Collision-Avoidance-8D16/README.md} \vspace{-4pt}